\documentclass{article}

 \usepackage[nonatbib, dblblindworkshop, final]{neurips_2025}
 \usepackage[numbers]{natbib}
 \usepackage{pifont}
\workshoptitle{Structured Probabilistic Inference \& Generative Modeling}

\usepackage[utf8]{inputenc} 
\usepackage[T1]{fontenc}    
\usepackage{hyperref}       
\usepackage{url}            
\usepackage{booktabs}       
\usepackage{amsfonts}       
\usepackage{nicefrac}       
\usepackage{microtype}      
\usepackage{xcolor}         
\usepackage{algorithm}
\usepackage{algorithmic}
\usepackage{pifont}
\usepackage{xcolor} 
\usepackage{graphicx}
\usepackage{placeins} 
\usepackage{amsmath,amssymb,amsthm}
\DeclareMathOperator*{\argmax}{arg\,max}

\title{\texttt{BP-Seg}: A graphical model approach to unsupervised and non-contiguous text segmentation using belief propagation}

\author{Fengyi Li, Kayhan Behdin, Natesh Pillai, Xiaofeng Wang, Zhipeng Wang, Ercan Yildiz  \\
LinkedIn Corporation \\ 
\texttt{\{fenli, kbehdin, napillai, xiaofwan, zhipwang, eyildiz\}@linkedin.com}}

\begin{document}

\maketitle

\begin{abstract}
Text segmentation based on the semantic meaning of sentences is a fundamental task with broad utility in many downstream applications. In this paper, we propose a graphical model-based unsupervised learning approach, named \texttt{BP-Seg} for efficient text segmentation. Our method not only considers local coherence, capturing the intuition that adjacent sentences are often more related, but also effectively groups sentences that are distant in the text yet semantically similar. This is achieved through belief propagation on the carefully constructed graphical models. Experimental results on both an illustrative example and a dataset with long-form documents demonstrate that our method performs favorably compared to competing approaches.
\end{abstract}

\section{Introduction}
Segmenting text into semantically coherent segments has been a long-studied problem in the field of natural language processing \citep{Pak2018TextST, Badjatiya2018AttentionBasedNT}. The applications of text segmentation range from information retrieval~\citep{Yu2023ImprovingLD, text_IR}, document summarization~\citep{cho-etal-2022-toward, Miculicich2023DocumentSW}, disclosure analysis~\citep{Wang2018TowardFA, song2021informational}, and optimizing prompts for large language models (LLMs) by extracting the most relevant parts~\citep{lewis2023rag, xu2025procutllmpromptcompression}. 

Traditional methods, whether supervised~\citep{Koshorek2018TextSA, Badjatiya2018AttentionBasedNT, Glavas2020TwoLevelTA} or unsupervised~\citep{TextTiling,graphseg,Barakat2020UnsupervisedDL}, mainly focus on \textit{contiguous} or \textit{sequential} text segmentation. The goal is to cluster \textit{consecutive} sentences in a way that ensures those within the same group are semantically more similar to each other than to sentences in different groups. For example, if a text consists of five sentences labeled $\{1,2,3,4,5\}$, traditional methods might segment them into groups such as $\{1,2\}$ and $\{3,4,5\}$. However, in practice, it is sometimes the case that $\{1,5\}$ are more semantically similar and should form one group, while $\{2,3,4\}$ form another. On the other hand, some frameworks for text segmentation disregard the adjacency relationships between sentences and the overall structure of the text. For example, methods such as $k$-means~\citep{lloyd1982least} treat each sentence (or its embedding) as an isolated data instance, without considering that a sentence is often more likely to be semantically connected to its adjacent sentences than to those that are farther apart. To the best of our knowledge, however, there is a lack of literature on semantic classification methods that take into account both adjacent and distant (non-adjacent) sentences. One example of such an application is prompt pruning for LLMs. When users write prompts, the sentences typically follow a logical flow, but some may be redundant. Splitting a prompt into groups, potentially non-sequential, while still accounting for the semantic coherence of contiguous sentences can facilitate downstream tasks such as prompt pruning, ultimately improving both efficiency and relevance in LLM interactions~\citep{SelfCP, xu2025procutllmpromptcompression}.




In this work, we propose a new framework for text segmentation that accounts for the fact that adjacent sentences are typically more related, while also enabling the grouping of non-contiguous sentences that are semantically similar. To achieve this, we first embed sentences into vector representations using sentence embeddings~\cite{sbert}, so that semantically similar sentences are mapped closer together in the embedding space. This allows us to form a graph from the text, where the nodes represent the embedded sentences and edges encode the strength of their semantic relationships.  We then apply Belief Propagation (BP, \citet{Pearl1982BP}), an inference algorithm used in graphical models, to generate clusters. To the best of our knowledge, this work presents the first successful application of BP to text segmentation that accounts for semantic meaning of sentences in both continuous and non-continuous settings.

\section{Method}\label{sec:method}
Our algorithm, named \texttt{BP-Seg}, consists of three main steps: sentence embedding, constructing the graphical model, and running BP. We discuss each of these steps in detail below.
\subsection{Sentence embeddings}
Given a text, represented as an ordered collection of sentences $\{S_i\}_{i=1}^n$, we can obtain their numerical vector representations using sentence embeddings. This can be efficiently achieved with libraries such as \texttt{transformers}, \texttt{sentence-transformers}, or \texttt{tensorflow\_hub}. Once encoded, semantically similar sentences are expected to have higher cosine similarity scores, indicating their closeness in the embedding space. We use $R_i$ to denote the sentence embedding of $S_i$.

\subsection{Constructing the graphical model}
The text segmentation process begins with the initialization of a set of cluster representatives, denoted as $C_j$, which serve as the reference representatives for segment assignments. In practice, these representatives are randomly selected from the set of input sentence embeddings. Given a text with $n$ sentences, we define $k$ clusters and randomly choose $k$ sentence embeddings as the initial representatives, $C = \{ C_1, C_2, \dots, C_k \}$, $C_j \in \{R_1, R_2, \dots, R_n\}$, with $C_i \neq C_j$. 
Let $x = \{ x_1, x_2, \dots, x_n \}$ be the segment assignments, where each $x_i$ represents the segment label assigned to sentence $S_i$. Therefore, each $x_i$ takes a discrete value from the set $\{1, 2, \dots, k\}$, where $k$ is the total number of segments.


Let $p_i(x_i)$ be the probability that the $i$th segment is assigned with a label $x_i$. If we assume the joint distribution factorizes, we can write $p(x_1, \ldots, x_n) = \frac{1}{Z}\prod_f \psi_f(x_f) \prod_g \psi_g(x_g)$, where $\psi_f$ represents unary factors and $\psi_g$ represents pairwise factors, and $Z$ is the normalizing constant. 
To be precise, $\psi_f(x_f)$ can be written in the form of $\psi_{i}(x_i)$, encoding how strongly the $i$th segment prefers the cluster $C_{x_i} \in C$. Similarly, the pairwise factors, $\psi_g(x_g)$ written in the form of $\psi_{i,j}(x_i,x_j)$, encode how compatible the $i$-th segment is assigned with the label $x_i$ and the $j$-th segment is assigned with the label $x_j$, where $x_i, x_j \in \{1,2,\cdots, k\}$. If $\psi_{i,j}(x_i,x_j)$ is large, 
it means that assigning the $i$-th segment with label $x_i$ and the $j$-th segment with label $x_j$ fits together well. In practice, one has the freedom to choose $\psi_f$ and $\psi_g$.  In this work, we set the node and edge factors as follows:
\begin{align} \label{eq:node_factor}
    \psi_i(x_i) = \exp(\mathrm{sim}(R_i, C_{x_i})),
\end{align}
and
\begin{align}\label{eq:edge_factor}
    \psi_{i,j}(x_i,x_j)=
\begin{cases}
1, & x_i=x_j\\
\exp(\lambda \left(\mathrm{sim}(R_i, R_j)-1\right)) & \mathrm{otherwise}
\end{cases}.
\end{align}
Here, $\mathrm{sim}(\cdot, \cdot)$ denotes the cosine similarity between two embeddings. A higher value of $\psi_i(x_i)$ indicates a greater likelihood that sentence $S_i$ belongs to segment $x_i$. Note also that $\psi_{i,j}(x_i, x_j) \leq 1$, with equality holding if and only if $x_i = x_j$, i.e., when the two sentences are assigned the same label. On the other hand, 
smaller values of $\psi_{i,j}$ reflect weaker semantic connections between sentences that differ in meaning.

In this work, we adopt specific forms of $\psi_i$ and $\psi_{i,j}$ to encode the semantic relationships between sentences and their assignments. However, in practice, one may choose alternative or domain-specific formulations, provided they are compatible with the desired inference algorithm.


\subsection{BP (Sum-Product) for text segmentation}

After assigning the node and edges with proper weights, we can start implementing the BP algorithm. The goal of BP is to maximize the marginal probability of segment assignments by iteratively exchanging messages between sentences and updating their segment beliefs. A message from $i$ to $j$ represents node $i$'s belief about the possible values of that node $j$ takes, considering all evidence except what comes from node $j$ itself.
At each iteration, every node (sentence embedding) $R_i$
sends a message to its neighboring node $R_j$, conveying how strongly $R_j$ is associated with a given segment. These messages incorporate both the unary factor, which measures the semantic similarity of a sentence to its assigned segment representative, and the pairwise factor, which enforces consistency between related sentences. Before process begins, the messages $m_{i \to j}(x_j)$ must be initialized. The simplest approach is to set all messages to be uniform, i.e., $m_{i \to j}(x_j) = 1/k$ for all $i, j$. This assumes no prior preference for any segment, allowing BP to refine the segmentation purely based on updates.
The message from $R_i$ to $R_j$ at iteration $t$ is updated as:
\begin{align} 
m_{i \to j}^{(t)}(x_j)
= \sum_{x_i}
\Big(
    \psi_i(x_i)\;\psi_{i,j}(x_i, x_j)  
    \prod_{k \in \{1, \ldots, n\}\setminus j}
    m_{k \to i}^{(t-1)}(x_i)
\Big),
\label{eq: message_passing}
\end{align}
where $\psi_i$ is the unary potential, and $\psi_{i,j}$ is the pairwise potential. Each node updates its belief about its segment assignment by accumulating incoming messages from all neighboring sentences:
\begin{align} \label{eq: BP_update}
   b_i(x_i)
\;\;\propto\;\;
\psi_{i}(x_i)
\prod_{j \in \{1, \ldots, n\}\setminus i}
m_{j \to i}(x_i).
\end{align}
This iterative process continues until convergence, where the segment labels stabilize. The final segmentation is determined by selecting the segment with the highest belief for each node,
\begin{align} \label{eq: assignment}
    x_i^* = \argmax_{x_i \in \{1,\cdots, k\}} b_i(x_i).
\end{align}
We summarize our proposed algorithm in Algorithm~\ref{alg:bp_text_segmentation}. More analysis of BP can be read in~\citet{murphy2013loopy,yedidia2003understanding}.

\begin{algorithm}
\caption{\texttt{BP-Seg}}
\label{alg:bp_text_segmentation}
\begin{algorithmic}[1]
    \STATE \textbf{Input:} Sentence embeddings $\{R_1, \dots, R_n\}$
    \STATE \textbf{Output:} Segment assignment $\{x_1^*, \dots, x_n^*\}$

    \STATE \textbf{Initialization:} Initialize $k$ segment representatives $\{C_1, \dots, C_k\}$. Initialize node and edge factors following~\eqref{eq:node_factor} and ~\eqref{eq:edge_factor}, and initialize all messages $m^{(0)}_{i \to j}(x_j) = 1/k$.
    \FOR{$t = 1$ to $T$}
        \FOR{each embedding $R_i$}
            \FOR{each $R_j \in \{1, \ldots, n\}\setminus i$}
                \STATE Update messages using~\eqref{eq: message_passing}.
            \ENDFOR
        \ENDFOR

    \ENDFOR
     \FOR{each embedding $R_i$}
            \STATE Update belief using~\eqref{eq: BP_update}.
    \ENDFOR
    
    \STATE \textbf{Final Segmentation:}
    \FOR{each embedding $R_i$}
        \STATE Assign segment with highest belief using~\eqref{eq: assignment}.
    \ENDFOR

    \RETURN Segment assignments $\{x_1^*, \dots, x_n^*\}$
\end{algorithmic}
\end{algorithm}

\section{Related work}\label{sec:related_work}
A prior work has explored a variation of using graph-based models for text segmentation. \texttt{GraphSeg}, proposed by~\citet{graphseg}, for example, also employs unsupervised learning for text segmentation within a graph-based framework. However, their primary objective is to produce \emph{contiguous} segmentations, whereas our method allows for a \emph{non-contiguous} segmentation that accounts for both neighboring and distant sentences. Moreover, their algorithm requires additional information, such as each word’s information content based on its relative frequency, whereas \texttt{BP-seg} relies \emph{solely} on embeddings and no external data. Moreover, after encoding sentences into embeddings and computing cosine similarities, their approach discovers segmentations by finding maximal cliques --- fundamentally different from our \emph{probabilistic} strategy, in which we seek an assignment that maximizes the marginal distribution. One could in principle apply $k$-means to group sentence embeddings; however, such a method is entirely context-agnostic and considers only pairwise embedding similarities.

\begin{table*}[t]
\centering
\footnotesize
\begin{tabular}{lcccccccccccc}
\hline
 & \multicolumn{2}{c}{\textbf{3--5}} & \multicolumn{2}{c}{\textbf{6--8}} & \multicolumn{2}{c}{\textbf{9--11}} & \multicolumn{2}{c}{\textbf{3--11}} & \multicolumn{2}{c}{\textbf{3--15}} & \multicolumn{2}{c}{\textbf{12--15}}\\
 & ARI & NMI & ARI & NMI & ARI & NMI & ARI & NMI  & ARI & NMI & ARI & NMI\\
\hline

\texttt{BP-Seg}   & 0.58 & 0.83 & \textbf{0.76} & \textbf{0.89} & \textbf{0.73} & \textbf{0.87} & \textbf{0.73} & \textbf{0.87} & \textbf{0.65} & \textbf{0.84} & \textbf{0.62} & \textbf{0.83} \\
\texttt{GraphSeg}   & \textbf{0.65} & \textbf{0.87} & 0.58 & 0.83 & 0.52 & 0.81 & 0.55 & 0.83 & 0.46 & 0.79 & 0.40 & 0.77 \\
$k$-means & 0.53 & 0.84 & 0.52 & 0.79 & 0.52 & 0.76 & 0.50 & 0.79 & 0.45 & 0.74 & 0.45 & 0.70 \\
\hline
\end{tabular}
\caption{Average performance on the Choi dataset measured using ARI and NMI across different subsets. Higher values indicate better performance. Note that a random segmentation method achieves a $0$ in ARI.}
\label{tab:choi_dataset_mean}
\end{table*}
\section{Experiments}\label{sec:experiments}
\subsection{The Illustrative Example}\label{sec:ill_ex}

To demonstrate the effectiveness of our proposed segmentation method, we compare its performance against \texttt{GraphSeg} (implemented using the code available here\footnote{\url{https://github.com/Dobatymo/graphseg-python}}), $k$-means, and a large language model (LLM). The input text used for segmentation, generated by GPT-4o with additional human-written content, is as follows:

\texttt{
The sun was shining brightly.
It was a beautiful morning.
I decided to go for a walk.
Suddenly, dark clouds appeared.
\textcolor{red}{I'll play tennis tomorrow.}
What are you doing?
Thunder rumbled in the distance.
The rain poured down heavily.
People ran for shelter.
\textcolor{red}{US Open is a tennis tournament.}
I am here working on my project.
The sun came out again.
\textcolor{red}{Who is going to win the US Open?}
}

Each sentence in the document is treated as an individual unit for segmentation. We evaluate the segmentation outcomes generated by our proposed method, \texttt{BP-Seg}, in comparison with three baselines: \texttt{GraphSeg}, $k$-means clustering, and a LLM (GPT-4o). To qualitatively assess the results, we focus on two key criteria:
(1) whether the method identifies more than one meaningful segment (i.e., produces more than one cluster), and
(2) whether the three semantically related tennis sentences, highlighted in red, are grouped together in the same cluster.
These criteria help determine both the model’s ability to detect structure and its sensitivity to semantic coherence. The full experimental setup, including parameter choices and additional visualizations across methods and configurations, can be found in Appendix~\ref{app:ill_example}.

For \texttt{BP-Seg} and $k$-means, we set the number of segments to $k = {2, 3, \cdots, 7}$. For the LLM, we explicitly prompt it to generate $k = {2, 3, \cdots, 7}$ segments. The prompt can be read in~\ref{app:llm_seg_prompt}. As observed, all methods generate more than one clusters. Furthermore, $k$-means successfully group sentences based on thematic coherence across all values of $k$. Both \texttt{BP-Seg} and the LLM successfully group the tennis-related sentences in the same cluster in $5$ out of the $6$ tested values of $k$. For \texttt{GraphSeg}, we set the minimum number of sentences per segment to 1 to allow maximum flexibility. However, despite trying various thresholds $\tau$, \texttt{GraphSeg} fails to cluster semantically related sentences effectively, as it is designed for contiguous segmentation. We summarize the results in Table~\ref{tab:short_clusters}.

\begin{table*}[t]
\centering
\footnotesize
\begin{tabular}{lcccccc}
\toprule
 &2 Clusters & 3 Clusters & 4 Clusters & 5 Clusters & 6 Clusters & 7 Clusters \\
\midrule
\texttt{BP-Seg}  & \ding{51}& \ding{51} & \ding{51} & \ding{55} & \ding{51} & \ding{51} \\
$k$-means  &\ding{51} & \ding{51}  & \ding{51}  & \ding{51}  & \ding{51}  & \ding{51}  \\
LLM  &\ding{51}& \ding{51} & \ding{51} & \ding{51} & \ding{55}  & \ding{51} \\
\texttt{Graph-Seg}  & \ding{55}& \ding{55} & \ding{55} & \ding{55} & \ding{55} & \ding{55} \\
\bottomrule
\end{tabular}
\caption{Performance on an illustrative example. Both \texttt{BP-Seg} and $k$-means are able to group tennis-related sentences into the same clusters, with varying numbers of clusters.}
\label{tab:short_clusters}
\end{table*}

\subsection{Choi dataset}\label{sec:choi_ex}
In this example, we implement our approach, \texttt{BP-Seg}, along with \texttt{GraphSeg} and $k$-means on the Choi dataset~\citep{choi2000advances}. The performance of the LLM is not included in this case, as its output does not include every sentence from the original text. Traditionally, for contiguous text segmentation, two evaluation metrics are commonly reported: $P_k$~\citep{beeferman1999statistical} and WindowDiff (WD)~\citep{pevzner-hearst-2002-critique}. The $P_k$ metric checks whether the boundary status (i.e., whether two sentences within a fixed-size window belong to the same segment) matches between the ground truth and the prediction. WD, on the other hand, measures whether the number of boundaries within the window is consistent with the ground truth. However, both metrics assume contiguity and are not suitable for evaluating non-contiguous text segmentation.

Therefore, we report Adjusted Rand Index (ARI)~\citep{hubert1985comparing, wagner2007comparing} and Normalized Mutual Information (NMI)~\citep{kvaalseth2017normalized}, which are appropriate for clustering-based evaluations. This adjustment accounts for the fact that the outputs of \texttt{BP-Seg} and $k$-means may result in non-contiguous segmentations, even though the ground truth segmentation is contiguous. Additionally, for efficiency, we use a variant of \texttt{BP-Seg} that is also based on message passing but offers faster computation. Please refer to Algorithm~\ref{alg:fast_BP} for more details.

The ARI and Normalized NMI are two widely used metrics for evaluating clustering and segmentation quality. ARI ranges from $-1$ to $1$, where a score of $1$ indicates a perfect agreement between the predicted segmentation and the ground truth, $0$ reflects a performance equivalent to random labeling, and negative values suggest an agreement worse than random chance. This makes ARI particularly informative in distinguishing between meaningful segmentations and those produced by chance. NMI, on the other hand, measures the mutual dependence between the predicted and true labels. It ranges from $0$ to $1$, with $1$ indicating complete alignment between the predicted and true segmentations, and $0$ representing statistical independence. Unlike ARI, NMI is insensitive to permutations of cluster labels, making it a complementary metric for evaluating clustering performance.

Table~\ref{tab:choi_dataset_mean} reports the mean segmentation performance of the three methods on the Choi dataset. As illustrated in the table, \texttt{BP-Seg} consistently outperforms the baseline methods across nearly all configurations, except in cases where each segment contains very few sentences (e.g., 3–5). This highlights the strength of our belief propagation-based formulation in capturing both local and global textual coherence. The standard deviation of the performance can be read in Table~\ref{tab:choi_dataset_std}.

\section{Conclusion}\label{sec:conclusion}
We presented \texttt{BP-Seg}, an efficient unsupervised approach for text segmentation using belief propagation. Our method effectively balances local contextual coherence with global semantic similarity, enabling more meaningful and flexible segmentation of text. Although designed for non-contiguous segmentation, experimental results show that \texttt{BP-Seg} outperforms several competitive methods on the standard contiguous segmentation task, achieving strong performance on metrics such as ARI and NMI. Looking forward, we aim to explore the utility of \texttt{BP-Seg} in real-world downstream applications. These include prompt pruning for LLMs, where segmenting prompts into semantically coherent chunks can lead to more efficient inference; information retrieval, where flexible segmentation can support better indexing and matching; and question answering, where isolating relevant spans is crucial for interpretability and performance. We believe the probabilistic and modular nature of \texttt{BP-Seg} offers a promising foundation for broader integration into these complex language understanding pipelines.

\section*{Limitations}
In this study, all examples are in English. The example in Section~\ref{sec:ill_ex} was generated by GPT-4o with additional human-written content, and the Choi dataset in Section~\ref{sec:choi_ex} is also synthetic. As a result, these examples may not accurately reflect real-world scenarios, and our evaluations are limited to these two cases. Nevertheless, we believe the insights from our findings will inspire further research in text segmentation and benefit a wide range of related applications.

\bibliographystyle{plainnat}
\bibliography{custom}
\newpage
\appendix

\section{Appendix}\label{sec:appendix}
\subsection{The Illustrave Example} \label{app:ill_example}
\subsubsection{Results using \texttt{BP-Seg}}
\textbf{Set $\lambda = 0.12$
in~\eqref{eq:edge_factor}}\\
\textbf{$k = 2$ segments}:
\texttt{
[Segment 1]:
The sun was shining brightly. It was a beautiful morning. I decided to go for a walk. Suddenly, dark clouds appeared. \textcolor{red}{I'll play tennis tomorrow.} What are you doing? The rain poured down heavily. People ran for shelter. \textcolor{red}{US Open is a tennis tournament.} I am here working on my project. The sun came out again. \textcolor{red}{Who is going to win the US Open?}
[Segment 2]:
Thunder rumbled in the distance. 
}
\\\\
\textbf{$k = 3$ segments}:
\texttt{
[Segment 1]:
The sun was shining brightly. It was a beautiful morning. I decided to go for a walk. Suddenly, dark clouds appeared. Thunder rumbled in the distance. The rain poured down heavily. The sun came out again.
[Segment 2]:
\textcolor{red}{I'll play tennis tomorrow.} What are you doing? \textcolor{red}{US Open is a tennis tournament.} I am here working on my project. \textcolor{red}{Who is going to win the US Open?}
[Segment 3]:
People ran for shelter.
}
\\\\
\textbf{$k = 4$ segments}:
\texttt{
[Segment 1]:
The sun was shining brightly. It was a beautiful morning. I decided to go for a walk. What are you doing? People ran for shelter. I am here working on my project. The sun came out again.
[Segment 2]:
Suddenly, dark clouds appeared. The rain poured down heavily.
[Segment 3]:
Thunder rumbled in the distance.
[Segment 4]:
\textcolor{red}{I'll play tennis tomorrow.} \textcolor{red}{US Open is a tennis tournament.} \textcolor{red}{Who is going to win the US Open?}
}
\\\\
\textbf{$k = 5$ segments}:
\texttt{
[Segment 1]:
The sun was shining brightly. It was a beautiful morning. Suddenly, dark clouds appeared. Thunder rumbled in the distance. People ran for shelter. \textcolor{red}{US Open is a tennis tournament.} The sun came out again.
[Segment 2]:
I decided to go for a walk.
[Segment 3]:
\textcolor{red}{I'll play tennis tomorrow.} What are you doing? \textcolor{red}{Who is going to win the US Open?}
[Segment 4]:
The rain poured down heavily.
[Segment 5]:
I am here working on my project.
}
\\\\
\textbf{$k = 6$ segments}:
\texttt{
[Segment 1]:
The sun was shining brightly. It was a beautiful morning. The rain poured down heavily. The sun came out again.
[Segment 2]:
I decided to go for a walk.
[Segment 3]:
\textcolor{red}{I'll play tennis tomorrow.} \textcolor{red}{US Open is a tennis tournament.} \textcolor{red}{Who is going to win the US Open?}
[Segment 4]:
Suddenly, dark clouds appeared. Thunder rumbled in the distance. 
[Segment 5]:
People ran for shelter.
[Segment 6]:
What are you doing? I am here working on my project.
}
\\\\
\textbf{$k = 7$ segments}:
\texttt{
[Segment 1]:
The sun was shining brightly. People ran for shelter.
[Segment 2]:
It was a beautiful morning. I decided to go for a walk.
[Segment 3]:
Suddenly, dark clouds appeared. Thunder rumbled in the distance.
[Segment 4]:
What are you doing? I am here working on my project.
[Segment 5]:
The rain poured down heavily.
[Segment 6]:
\textcolor{red}{I'll play tennis tomorrow.} \textcolor{red}{US Open is a tennis tournament.} \textcolor{red}{Who is going to win the US Open?}
[Segment 7]:
The sun came out again.
}
\subsubsection{Results using \texttt{GraphSeg}}

\textbf{treshold $\tau = 0.1, 0.3, 0.5, 0.7, 0.9$ and minimal segment size $n=1$}:
\texttt{
[Segment 1]:
The sun was shining brightly. It was a beautiful morning.
[Segment 2]:
I decided to go for a walk. Suddenly, dark clouds appeared. What are you doing?
[Segment 3]:
Thunder rumbled in the distance. The rain poured down heavily.
[Segment 4]:
People ran for shelter. \textcolor{red}{US Open is a tennis tournament.}
[Segment 5]:
I am here working on my project. The sun came out again. \textcolor{red}{Who is going to win the US Open?}
}

\subsubsection{Results using $k$-means}
\textbf{$k = 2$ segments}:\texttt{
[Segment 1]:
\textcolor{red}{I'll play tennis tomorrow.} What are you doing? \textcolor{red}{US Open is a tennis tournament.} I am here working on my project. \textcolor{red}{Who is going to win the US Open?}
[Segment 2]:
The sun was shining brightly. It was a beautiful morning. I decided to go for a walk. Suddenly, dark clouds appeared. Thunder rumbled in the distance. The rain poured down heavily. People ran for shelter. The sun came out again.
}
\\\\
\textbf{$k = 3$ segments}:\texttt{
[Segment 1]:
The sun was shining brightly. It was a beautiful morning. Suddenly, dark clouds appeared. Thunder rumbled in the distance. The rain poured down heavily. People ran for shelter. The sun came out again.
[Segment 2]:
\textcolor{red}{I'll play tennis tomorrow.} \textcolor{red}{US Open is a tennis tournament.} \textcolor{red}{Who is going to win the US Open?}
[Segment 3]:
I decided to go for a walk. What are you doing? I am here working on my project.
}
\\\\
\textbf{$k = 4$ segments}:\texttt{
[Segment 1]:
The sun was shining brightly. It was a beautiful morning. Suddenly, dark clouds appeared. Thunder rumbled in the distance. The rain poured down heavily. The sun came out again.
[Segment 2]:
\textcolor{red}{I'll play tennis tomorrow.} \textcolor{red}{US Open is a tennis tournament.} \textcolor{red}{Who is going to win the US Open?}
[Segment 3]:
I decided to go for a walk. What are you doing? I am here working on my project.
[Segment 4]:
People ran for shelter.
}
\\\\
\textbf{$k = 5$ segments}:\texttt{
[Segment 1]:
I decided to go for a walk. People ran for shelter.
[Segment 2]:
The sun was shining brightly. Suddenly, dark clouds appeared. Thunder rumbled in the distance. The sun came out again.
[Segment 3]:
What are you doing? I am here working on my project.
[Segment 4]:
\textcolor{red}{I'll play tennis tomorrow.} \textcolor{red}{US Open is a tennis tournament.} \textcolor{red}{Who is going to win the US Open?}
[Segment 5]:
It was a beautiful morning. The rain poured down heavily.
}
\\\\
\textbf{$k = 6$ segments}:\texttt{
[Segment 1]:
Suddenly, dark clouds appeared. Thunder rumbled in the distance. The rain poured down heavily.
[Segment 2]:
What are you doing? I am here working on my project.
[Segment 3]:
\textcolor{red}{I'll play tennis tomorrow.} \textcolor{red}{US Open is a tennis tournament.} \textcolor{red}{Who is going to win the US Open?}
[Segment 4]:
The sun was shining brightly. The sun came out again.
[Segment 5]:
People ran for shelter.
[Segment 6]:
It was a beautiful morning. I decided to go for a walk.
}
\\\\
\textbf{$k = 7$ segments}:\texttt{
[Segment 1]:
I am here working on my project.
[Segment 2]:
The sun was shining brightly. It was a beautiful morning. The sun came out again.
[Segment 3]:
\textcolor{red}{I'll play tennis tomorrow.} \textcolor{red}{US Open is a tennis tournament.} \textcolor{red}{Who is going to win the US Open?}
[Segment 4]:
I decided to go for a walk. What are you doing?
[Segment 5]:
People ran for shelter.
[Segment 6]:
Suddenly, dark clouds appeared. Thunder rumbled in the distance.
[Segment 7]:
The rain poured down heavily.
}

\subsubsection{Results using LLM (GPT-4o)}\label{app:llm_seg_prompt}
\textbf{Prompt}:
\texttt{
Please generate \{number\_seg\} segmentation for the following text according to their sematic meaning. Adjacent sentences do not have to be in the same segment.
The sun was shining brightly.
It was a beautiful morning.
I decided to go for a walk.
Suddenly, dark clouds appeared.
I'll play tennis tomorrow.
What are you doing?
Thunder rumbled in the distance.
The rain poured down heavily.
People ran for shelter.
US Open is a tennis tournament.
I am here working on my project.
The sun came out again.
Who is going to win the US Open?
Return the output in the following format:
[Segment 1]: The sun was shining brightly. It was a beautiful morning.
[Segment 2]: I decided to go for a walk. Suddenly, dark clouds appeared.
...
}
\\\\
\textbf{$k = 2$ segments}:\texttt{ [Segment 1]: The sun was shining brightly. It was a beautiful morning. Suddenly, dark clouds appeared. Thunder rumbled in the distance. The rain poured down heavily. People ran for shelter. The sun came out again. [Segment 2]: I decided to go for a walk. \textcolor{red}{I'll play tennis tomorrow.} What are you doing? \textcolor{red}{US Open is a tennis tournament.} I am here working on my project. \textcolor{red}{Who is going to win the US Open?}
}
\\\\
\textbf{$k = 3$ segments}:\texttt{ [Segment 1]: The sun was shining brightly. It was a beautiful morning. The sun came out again. [Segment 2]: I decided to go for a walk. Suddenly, dark clouds appeared. Thunder rumbled in the distance. The rain poured down heavily. People ran for shelter. [Segment 3]: \textcolor{red}{I'll play tennis tomorrow.} \textcolor{red}{US Open is a tennis tournament.} \textcolor{red}{Who is going to win the US Open?}
}
\\\\
\textbf{$k = 4$ segments}:\texttt{ [Segment 1]: The sun was shining brightly. It was a beautiful morning. The sun came out again. [Segment 2]: I decided to go for a walk. Suddenly, dark clouds appeared. Thunder rumbled in the distance. The rain poured down heavily. People ran for shelter. [Segment 3]: \textcolor{red}{I'll play tennis tomorrow.} \textcolor{red}{US Open is a tennis tournament.} \textcolor{red}{Who is going to win the US Open?} [Segment 4]: What are you doing? I am here working on my project.
}
\\\\
\textbf{$k = 5$ segments}:\texttt{ [Segment 1]: The sun was shining brightly. It was a beautiful morning. The sun came out again. [Segment 2]: I decided to go for a walk. Suddenly, dark clouds appeared. Thunder rumbled in the distance. The rain poured down heavily. People ran for shelter. [Segment 3]: \textcolor{red}{I'll play tennis tomorrow.} \textcolor{red}{US Open is a tennis tournament.} \textcolor{red}{Who is going to win the US Open?} [Segment 4]: What are you doing? [Segment 5]: I am here working on my project.
}
\\\\
\textbf{$k = 6$ segments}:\texttt{ [Segment 1]: The sun was shining brightly. It was a beautiful morning. [Segment 2]: I decided to go for a walk. Suddenly, dark clouds appeared. Thunder rumbled in the distance. The rain poured down heavily. People ran for shelter. The sun came out again.  
[Segment 3]: \textcolor{red}{I'll play tennis tomorrow.}  
[Segment 4]: What are you doing?  
[Segment 5]: \textcolor{red}{US Open is a tennis tournament.} \textcolor{red}{Who is going to win the US Open?}  
[Segment 6]: I am here working on my project.  
}
\\\\
\textbf{$k = 7$ segments}:\texttt{
[Segment 1]: The sun was shining brightly. It was a beautiful morning.[Segment 2]: I decided to go for a walk. Suddenly, dark clouds appeared. [Segment 3]: Thunder rumbled in the distance. The rain poured down heavily. People ran for shelter. [Segment 4]: The sun came out again. [Segment 5]: \textcolor{red}{\textcolor{red}{I'll play tennis tomorrow.}} US Open is a tennis tournament. \textcolor{red}{\textcolor{red}{Who is going to win the US Open?}} [Segment 6]: What are you doing? [Segment 7]: I am here working on my project.
}

\subsection{Choi dataset}\label{app:choi}
Here $m_i(x_i)$ can be viewed as a running preference or partial belief for node $i$ in label $x_i$. We also set $\psi_i(x_i) = \mathrm{sim}(R_i, C_{x_i})$ and $\psi_{i,j}= \lambda\mathrm{sim}(R_i, R_j)\exp(-\vert i-j\vert^2/\sigma)$. 
For comparison, we set the number of initial clusters in \texttt{BP-Seg} to be equal to the number of sentences, although many clusters may become empty after the algorithm terminates. We fix the number of iterations to $T = 5$, and choose $\sigma = 10$ and $\lambda = 300$ to account for the length of the text. For \texttt{GraphSeg}, we set the threshold parameter to $\tau = 0.2$ and the minimum segment size to $n = 1$ to allow maximum flexibility. For the $k$-means baseline, we cap the number of clusters $k$ at 20 and rely on the default number of iterations in \texttt{sklearn.cluster.KMeans}.

\onecolumn

\begin{algorithm}[H]
\caption{Fast \texttt{BP-Seg}}
\label{alg:fast_BP}
\begin{algorithmic}[1]
    \STATE \textbf{Input:} Sentence embeddings $\{R_1, \dots, R_n\}$
    \STATE \textbf{Output:} Segment assignment $\{x_1^*, \dots, x_n^*\}$

    \STATE \textbf{Initialization:} Initialize $k$ segment representatives $\{C_1, \dots, C_k\}$. Initialize node and edge factors following~\eqref{eq:node_factor} and ~\eqref{eq:edge_factor}, and initialize all messages $m^{(0)}_{i}(x_i) = 1/k$.
    \FOR{$t = 1$ to $T$}
        \FOR{each embedding $R_i$}
            \FOR{$x_i = \{1,2,\cdots, k\}$}
                \STATE $m_i^{(t)}(x_i) = \psi_{i}(x_i) + \sum_{j=1}^n \psi_{i,j} m_j^{(t-1)}(x_i)$
            \ENDFOR
        \ENDFOR

    \ENDFOR
     \FOR{each embedding $R_i$}
            \STATE Update belief using $b_i(x_i) = \psi_i(x_i) + m_i^{(T)}(x_i)$
    \ENDFOR
    
    \STATE \textbf{Final Segmentation:}
    \FOR{each embedding $R_i$}
        \STATE Assign segment using $x_i^* = \argmax_{x_i \in \{1,\cdots, k\}} b_i(x_i)$.
    \ENDFOR

    \RETURN Segment assignments $\{x_1^*, \dots, x_n^*\}$
\end{algorithmic}
\end{algorithm}

\begin{table}[H]
\centering
\footnotesize
\begin{tabular}{lcccccccccccc}
\hline
 & \multicolumn{2}{c}{\textbf{3--5}} & \multicolumn{2}{c}{\textbf{6--8}} & \multicolumn{2}{c}{\textbf{9--11}} & \multicolumn{2}{c}{\textbf{3--11}} & \multicolumn{2}{c}{\textbf{3--15}} & \multicolumn{2}{c}{\textbf{12--15}}\\
 & ARI & NMI & ARI & NMI & ARI & NMI & ARI & NMI  & ARI & NMI & ARI & NMI\\
\hline
\texttt{BP-Seg}  & 0.10 & 0.04 & 0.08 & 0.04 & 0.08 & 0.04 & 0.09 & 0.04 & 0.08 & 0.03 & 0.08 & 0.03 \\
\texttt{GraphSeg}  & 0.11 & 0.04 & 0.07 & 0.03 & 0.06 & 0.02 & 0.08 & 0.03 & 0.06 & 0.02 & 0.07 & 0.02 \\
$k$-means & 0.09 & 0.03 & 0.08 & 0.04 & 0.08 & 0.05 & 0.08 & 0.04 & 0.07 & 0.04 & 0.07 & 0.05 \\
\hline
\end{tabular}
\caption{Standard deviation on the Choi dataset measured using ARI and NMI across different subsets.}
\label{tab:choi_dataset_std}
\end{table}
\end{document}